\documentclass[runningheads]{llncs}

\usepackage{cite}
\usepackage{amsmath,amssymb,amsfonts}
\usepackage{algorithmic}
\usepackage{algorithm}
\usepackage{graphicx}
\usepackage{textcomp}
\usepackage{xcolor}
\usepackage{booktabs}
\usepackage{multirow}
\usepackage{subcaption}
\usepackage{hyperref}
\usepackage{balance}
\usepackage{enumitem}
\usepackage{xspace}
\usepackage{tikz}
\usepackage{pgfplots}
\pgfplotsset{compat=1.18}
\usepackage[table,xcdraw]{xcolor}
\usepackage[most]{tcolorbox}

\newtcolorbox{myquote}[1][]{%
    colback=black!5,
    colframe=black!5,
    notitle,
    sharp corners,
    borderline west={2pt}{0pt}{gray!80!black},
    enhanced,
    breakable,
    }

\newcommand{\sys}{Vanguard\xspace}

\graphicspath{{figures/}}

\begin{document}


\title{Load Testing for Machine Learning Model Serving Systems at Scale}

\author{Amr S. Abdelfattah\inst{1}\orcidID{0000-0001-7702-0059} \and
Nakul Tirumalai\inst{1} \and
Indu Mohanan\inst{1} \and
Xiao Li\inst{1} \and
Pengchao Wang\inst{1} 
Dinakar Dhurjati\inst{1} \and
Eric Sung\inst{1}}
\authorrunning{Amr S. Abdelfattah et al.}
\institute{Meta Inc., Menlo Park, USA\\
\email{\{amrs, nakult, imohanan, ilx, wpc, dinakar, ericsung\}@meta.com}}


\maketitle

\begin{abstract}
Machine learning (ML) model serving has become a dominant consumer of GPU infrastructure, yet capacity planning in these systems remains largely ad hoc. Under-provisioning leads to service-level objective (SLO) violations and production incidents, while over-provisioning results in substantial resource waste. This paper presents \sys, an industrial load testing framework for ML serving systems that systematically estimates serving capacity through an adaptive, feedback-driven search strategy. The approach leverages real-time performance signals, incorporating dampening, spike tolerance, and convergence detection to efficiently identify maximum sustainable throughput under SLO constraints. We evaluate \sys through a longitudinal analysis of 14 industrial case studies spanning four ML architecture classes: recommendation, ranking, vision, and NLP. This study demonstrates that systematic load testing leads to substantial improvements in GPU resource efficiency and operational reliability. Prior to adopting \sys, a significant fraction of model launches were under-provisioned, resulting in recurring incidents; these issues were substantially reduced after deployment. Our results show that ML-specific design decisions are critical to accurate capacity estimation: workload calibration using recorded traffic reduces estimation error from approximately 30\% to 2--6\%, while proper warmup handling yields a 22.2\% improvement in accuracy. Further analysis reveals key factors influencing prediction error, including model size and co-location effects. This paper distills six lessons and derive architectural guidelines for ML load testing, offering actionable insights for building reliable and efficient ML serving systems.


\end{abstract}


\section{Introduction}\label{sec:introduction}

Machine learning inference serving has become one of the largest consumers of GPU infrastructure at technology companies~\cite{hazelwood2018applied, gupta2020architectural}.
Recommendation and ranking models alone require thousands of GPU cards to serve billions of requests daily.
As organizations deploy increasingly complex models---from sparse embedding-based recommenders to dense transformer architectures---the impact of \emph{incorrect capacity provisioning} has grown proportionally.

\noindent\textbf{Problem Statement.} Under-provisioned models violate service-level objectives (SLOs), causing degraded user experience and incidents.
On the other hand, over-provisioned models waste GPU resources that could serve other workloads or reduce operational requirements.
Despite the maturity of load testing for traditional web services~\cite{jiang2015loadtesting}, ML model serving presents unique challenges that existing tools do not adequately address:

\begin{itemize}[leftmargin=*,nosep]
 \item \textit{Model warmup.} GPU-based inference requires JIT (Just-In-Time) compilation, CUDA kernel caching~\cite{stratton2008mcuda}, and model weight loading, creating transient performance that does not reflect steady state.
 \item \textit{Dynamic batching.} Modern serving systems batch incoming requests for throughput efficiency, making per-request latency dependent on concurrent load---a nonlinear relationship absent in traditional services.
 \item \textit{Architecture heterogeneity.} A single organization may serve models ranging from sub-millisecond embedding lookups to multi-second generative outputs, each requiring fundamentally different testing approaches.
 \item \textit{Hardware sensitivity.} Inference performance varies across GPU generations, memory configurations, and co-location patterns, making capacity estimates hardware-specific.
\end{itemize}

General-purpose load testing tools such as JMeter~\cite{jmeter}, Locust~\cite{locust}, and k6~\cite{k6} lack abstractions for these ML-specific concerns.
ML benchmarking suites such as MLPerf Inference~\cite{reddi2020mlperf} focus on standardized offline throughput measurement rather than realistic capacity estimation.
Neither category provides \emph{adaptive search strategies} that converge efficiently on a model's maximum sustainable throughput under realistic SLO constraints.

\noindent\textbf{Objective.} This paper presents a load-testing framework designed to efficiently serve large-scale industrial models, addressing the limitations of general-purpose tools in handling such demands. We present \sys, an industrial load testing framework that has been used to test hundreds of ML models.
\sys addresses the gap between traditional load testing and ML-specific capacity planning through an adaptive feedback-driven search strategy, a multi-metric health assessment engine, and automated test scheduling.

\noindent\textbf{Contributions.} This paper contributes:

\begin{enumerate}[leftmargin=*,nosep]
 \item \textit{Adaptive strategy.} A feedback-driven load testing approach with dampening, spike tolerance, and convergence detection, evaluated on 14 ML models across four architectures (Sections~\ref{sec:methodology}, \ref{sec:results}).
 \item \textit{Design and accuracy analysis.} An ablation and regression study quantifying ML-specific design choices (e.g., warmup, batching, GPU health) and identifying factors affecting capacity prediction accuracy (Section~\ref{sec:results}).
 \item \textit{Reproducibility and impact.} Validation across runs, hardware, and time, plus 14 industrial case studies showing GPU efficiency gains and incident prevention (Section~\ref{sec:results}).
 \item \textit{Lessons learned.} Six actionable insights for researchers and practitioners (Section~\ref{sec:discussion}).
\end{enumerate}



\noindent\textbf{Paper Organization.} The remainder of this paper is organized as follows.
Section~\ref{sec:system} describes the \sys system design.
Section~\ref{sec:methodology} details our experimental methodology.
Section~\ref{sec:results} presents results for each research question.
Section~\ref{sec:discussion} discusses lessons learned, implications, and threats to validity.
Section~\ref{sec:related} surveys related work, and Section~\ref{sec:conclusion} concludes.

\section{System Design and Methodology}\label{sec:system}

This section describes the architecture of \sys, its load-testing strategies with a focus on the adaptive feedback-driven search, and the health assessment engine. We emphasize the design choices that address ML-specific challenges, as well as the methodological decisions \sys employs to overcome them.




\vspace{-0.3cm}
\subsection{Architecture Overview}\label{sec:system:arch}
\begin{figure*}[h]
 \centering
 \includegraphics[width=0.95\linewidth]{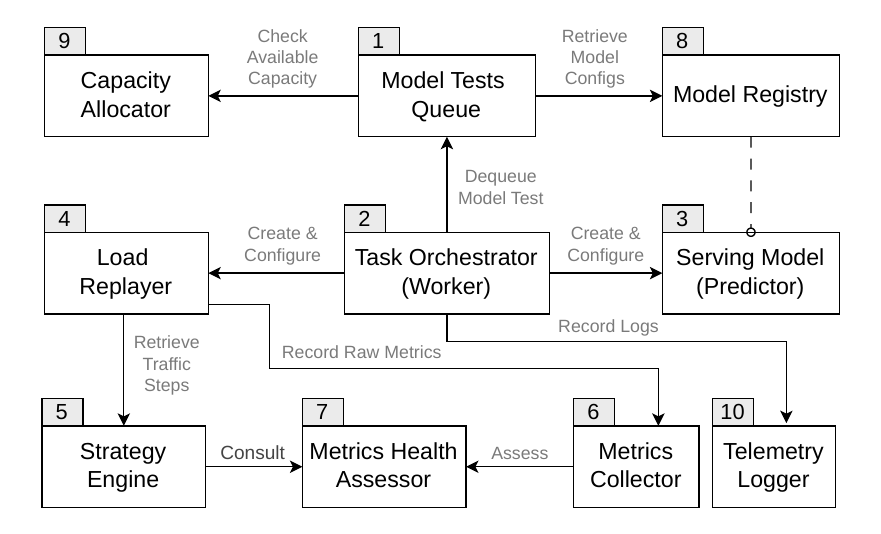}
 \caption{Architecture of \sys. Arrows indicate data flow, the Strategy Engine is pluggable across strategies (Section 2.3), and numbered components correspond to the explanation sequence in the text.}
 \label{fig:architecture}
\end{figure*}

\vspace{-0.1cm}
\sys is a distributed load testing system designed for ML model serving evaluation. Figure~\ref{fig:architecture} shows its architecture, which consists of ten components. Overall, these components collectively enable automated, scalable, and reproducible load testing for ML systems. Internal components of the framework include the Task Orchestrator, Replayer, Strategy Engine, Metrics Collector, Metrics Health Assessor, and Telemetry Logger, which together implement the execution, control, and evaluation logic. In contrast, components such as the Serving Model, Model Registry, Capacity Allocator, and Model Tests Queue are either external or infrastructure-facing, providing model artifacts, scheduling triggers, and computational resources required to drive the testing process. These components are described as follows:

\begin{enumerate}[leftmargin=*,nosep]

 \item[1.] \textbf{Model Tests Queue.} Provides a centralized queue that stores pending load-testing jobs with priority-aware scheduling. It ensures fair and efficient processing by allowing higher-priority models (e.g., high-priority or newly updated models) to be executed before lower-priority ones, while maintaining scalability and fault tolerance.

 \item[2.] \textbf{Task Orchestrator (a.k.a. Worker).} Manages the lifecycle of load test executions, including capacity allocation, service startup, test execution, and cleanup. It processes test cases from a distributed queue with atomic dequeue semantics to prevent race conditions between workers.

 \item[3.] \textbf{Serving Model (a.k.a. Predictor).} The deployed model inference service under test, responsible for handling incoming prediction requests. It encapsulates the model artifact, runtime environment (e.g., containerized GPU/CPU execution), and serving stack (e.g., REST/gRPC endpoints), exposing configurable parameters such as batching, concurrency limits, and hardware utilization policies.

 \item[4.] \textbf{Load Replayer.} Sends prediction requests to the model serving instance at a controlled rate. It supports smooth Query Per Second (QPS) transitions (linear ramp between target rates), configurable warmup periods, and open-loop request generation to avoid coordinated omission~\cite{tene2015coordinated}. Coordinated omission refers to a measurement bias where delays in processing suppress the generation of new requests, thereby masking true latency under load.

 \item[5.] \textbf{Strategy Engine.} Implements the strategy pattern~\cite{gamma1994design} to decouple the search algorithm from the orchestration logic. Each strategy implements a common interface (\texttt{next\_rate(metrics)} $\rightarrow$ \texttt{rate}) that returns the next QPS level to test based on observed metrics.

 \item[6.] \textbf{Metrics Collector.} Aggregates response metrics (latency percentiles, throughput, error rates, GPU utilization) over configurable time windows, providing both instantaneous and aggregate views to the strategy engine.

 \item[7.] \textbf{Metrics Health Assessor.} Evaluates collected metrics against per-model SLO definitions. It supports multi-metric constraints with sustained-violation requirements (e.g., ``P90 latency exceeds 50ms for 3 consecutive windows'') to reduce false positives from transient spikes.

 \item[8.] \textbf{Model Registry.} Maintains versioned metadata and artifacts for all models' snapshots, including model binaries, configuration parameters, SLO definitions, and serving specifications. It serves as the source of truth for retrieving model details required to instantiate reproducible load-testing environments.

 \item[9.] \textbf{Capacity Allocator.} Responsible for provisioning and managing compute resources (e.g., GPUs/CPUs) required for executing load tests. It supports both pre-allocated (static) capacity and on-demand (dynamic) allocation strategies, ensuring efficient utilization while meeting test isolation and performance requirements.

 \item[10.] \textbf{Telemetry Logger.} Records all test results, configuration parameters, and intermediate metrics to internal data warehouses for offline analysis, trend tracking, and reproducibility verification.

\end{enumerate}

From an architectural perspective, the workflow begins with the Task Orchestrator retrieves test jobs from the Model Tests Queue, consults the Model Registry for configuration details, and requests resources from the Capacity Allocator to deploy the Serving Model. Once deployed, the Replayer generates controlled traffic while the Metrics Collector and Health Assessor continuously evaluate system behavior. The Strategy Engine dynamically adjusts load levels based on observed performance, and the Telemetry Logger records all outcomes. This coordinated interaction enables systematic exploration of model performance under varying loads, ultimately achieving reliable and comprehensive load-testing outcomes.

\vspace{-0.3cm}
\subsection{Load Testing Strategies}\label{sec:system:strategies}
\vspace{-0.1cm}
\sys implements five load testing strategies, each optimizing for different trade-offs between accuracy, test duration, and robustness. We briefly describe the four baseline strategies before presenting the adaptive feedback-driven search in detail.
\vspace{-0.3cm}
\subsubsection{Baseline Strategies}

\sys supports four strategies beyond the adaptive approach:
(1)~\emph{Multi-Rate Sweep} tests a predetermined sequence of QPS levels from low load to expected saturation, each for a fixed duration, and estimates capacity as the highest level where all SLOs are satisfied---it is simple and predictable but its accuracy depends on the granularity of levels tested;
(2)~\emph{Binary Search} performs a binary search over the QPS space $[q_{\min}, q_{\max}]$, iteratively narrowing the range based on pass/fail results with retry logic for measurement noise, converging in $O(\log \frac{q_{\max} - q_{\min}}{\varepsilon})$ steps;
(3)~\emph{Direct Controller Integration} delegates QPS selection to an external capacity controller that implements the same autoscaling algorithm used in production-like environment, ensuring consistency between test-time and deployment-time decisions; and
(4)~\emph{Constant-Rate Stability Testing} holds a fixed QPS for an extended duration (1--4 hours) to assess system stability and detect regressions, rather than estimating capacity.
\vspace{-0.3cm}
\subsubsection{Adaptive Feedback-Driven Search}

We focus our detailed presentation on the adaptive strategy for three reasons. First, it is an effective technical contribution of this work, introducing real-time feedback mechanisms—damping, spike tolerance, and convergence detection—that are not present in most of the prior load testing approaches and are designed to better handle the nonlinear behavior of ML serving systems.
Second, unlike direct controller integration, the adaptive strategy is self-contained and portable across serving deployments without coupling to a specific autoscaling implementation. Third, the remaining strategies (sweep, binary search, constant-rate) are straightforward adaptations of well-known techniques, whereas the adaptive strategy introduces ML-specific mechanisms that warrant detailed exposition.

The adaptive strategy uses real-time metric feedback with three mechanisms: (1)~\emph{dampening}, which reduces step size as the search approaches the capacity boundary to avoid overshooting; (2)~\emph{spike tolerance}, which ignores transient metric spikes below a configurable threshold to prevent premature convergence; and (3)~\emph{convergence detection}, which monitors the variance of recent capacity estimates and declares convergence when the coefficient of variation falls below a threshold. Algorithm~\ref{alg:adaptive} presents the pseudocode.

\begin{algorithm}[h]
\caption{Adaptive Feedback-Driven Search Strategy}\label{alg:adaptive}
\scriptsize 
\begin{algorithmic}[1]
\REQUIRE Initial range $[q_{\min}, q_{\max}]$, dampening factor $\alpha$, spike tolerance $\tau$, convergence threshold $\epsilon$, max iterations $N$
\ENSURE Capacity estimate $q^*$
\STATE $q \leftarrow q_{\min}$; $\text{step} \leftarrow (q_{\max} - q_{\min}) / 4$
\STATE $\text{history} \leftarrow [\,]$; $\text{best} \leftarrow 0$
\FOR{$i = 1$ \TO $N$}
 \STATE $M \leftarrow \textsc{RunAtRate}(q, \text{duration})$
 \STATE $h \leftarrow \textsc{AssessHealth}(M)$
 \IF{$h = \textsc{Healthy}$}
 \STATE $\text{best} \leftarrow \max(\text{best}, q)$
 \STATE $q \leftarrow q + \text{step}$
 \ELSIF{$h = \textsc{Warn}$}
 \STATE $\text{best} \leftarrow \max(\text{best}, q)$
 \STATE $\text{step} \leftarrow \text{step} \times \alpha$ \COMMENT{Dampen}
 \STATE $q \leftarrow q + \text{step}$
 \ELSIF{$\textsc{SpikeCount}(M) \leq \tau$}
 \STATE $q \leftarrow q + \text{step} \times \alpha$ \COMMENT{Tolerate}
 \ELSE
 \STATE $q \leftarrow q - \text{step}$
 \STATE $\text{step} \leftarrow \text{step} \times \alpha$ \COMMENT{Backtrack}
 \ENDIF
 \STATE $\text{history}.\text{append}(\text{best})$
 \IF{$|\text{history}| \geq 3$ \AND $\textsc{CV}(\text{history}[-3:]) < \epsilon$}
 \RETURN $\text{best}$ \COMMENT{Converged}
 \ENDIF
\ENDFOR
\RETURN $\text{best}$
\end{algorithmic}
\end{algorithm}

The dampening factor $\alpha$ and spike tolerance $\tau$ must be tuned; our defaults ($\alpha = 0.5$, $\tau = 2$) work well for recommendation and ranking models but may need adjustment for other model types with higher variance.
\vspace{-0.4cm}
\subsection{Health Assessment Engine}\label{sec:system:health}
\vspace{-0.1cm}
The objective of the health assessment engine is to map raw metrics to five health states that classify the serving instance based on current metrics relative to SLO thresholds:

\begin{itemize}[leftmargin=*,nosep]
 \item \textsc{Healthy}: all metrics within SLO with margin ($\geq 20\%$ headroom).
 \item \textsc{Warn}: all metrics within SLO but approaching thresholds ($< 20\%$ headroom).
 \item \textsc{Unhealthy}: one or more metrics violating SLO intermittently.
 \item \textsc{Critical}: sustained SLO violation on primary metrics.
 \item \textsc{Overloaded}: system unable to process requests at the offered rate.
\end{itemize}

\begin{figure*}[h]
 \centering
 \includegraphics[width=\linewidth]{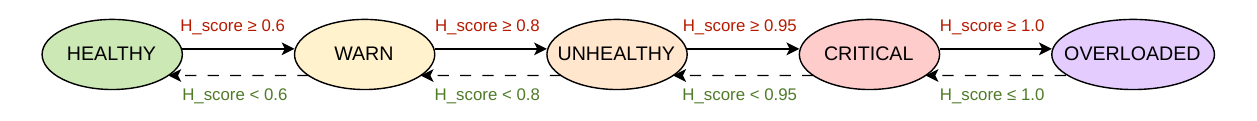}
 \caption{Health states: solid line = degrade; dashed line = recover.}
 \label{fig:health}
\end{figure*}

To improve assessment reliability, the health assessment engine incorporates two mechanisms, illustrated by the state machine in Figure~\ref{fig:health}.

\textbf{1. Multi-metric weighting.} Rather than evaluating each metric independently, the engine computes a weighted health score:
\begin{equation}\label{eq:health}
 H_{\text{score}} = \sum_{i=1}^{k} w_i \cdot \text{normalize}(m_i, t_i)
\end{equation}
where $w_i$ is the weight for metric $m_i$, $t_i$ is its SLO threshold, and $\text{normalize}(\cdot)$ maps the metric to $[0, 1]$ relative to its threshold. The health state is then determined by the score: $H_{\text{score}} < 0.6 \rightarrow \textsc{Healthy}$, $0.6$--$0.8 \rightarrow \textsc{Warn}$, $0.8$--$0.95 \rightarrow \textsc{Unhealthy}$, $0.95$--$1.0 \rightarrow \textsc{Critical}$, $> 1.0 \rightarrow \textsc{Overloaded}$.

\textbf{2. Hysteresis.} To prevent rapid oscillation between states (``flapping''), the engine requires a state to persist for $n$ consecutive measurement windows before transitioning. Upward transitions (toward \textsc{Overloaded}) use $n = 2$ windows for responsiveness, while downward transitions (toward \textsc{Healthy}) use $n = 3$ windows for stability. These window sizes were selected via empirical tuning to balance responsiveness and stability, as longer windows added latency without meaningful robustness gains. This asymmetry prioritizes fast degradation detection and cautious recovery confirmation.

\section{Experimental Methodology}\label{sec:methodology}

This section presents our research questions, experimental design, model selection, ground truth definition, and statistical analysis plan.
We investigate the following research questions (RQs):

\begin{itemize}[leftmargin=*]
 \item[\textit{RQ1}] \emph{How do ML-specific design decisions influence load testing accuracy compared to approaches that lack these features?} We measure the impact of warmup handling, batching awareness, and GPU health monitoring through ablation experiments.

 \item[\textit{RQ2}] \emph{How accurately do load test results predict serving capacity, and what factors explain prediction errors?} We analyze paired (estimate, actual) data for model launches and build a regression model of error factors.
\end{itemize}

\subsection{Experimental Design:}\label{sec:methodology:design}
\vspace{-0.1cm}
The experimental design outlines the selection of experimental variables, the criteria used for model selection, and the construction of ground truth used to support answering the research questions.


\begin{table*}[h]
\centering
\scriptsize
\caption{Experimental Variables}\label{tab:variables}
\begin{tabular}{@{}llp{6.5cm}@{}}
\toprule
\textbf{Type} & \textbf{Variable} & \textbf{Levels/Range} \\
\midrule
\multirow{3}{*}{Independent} & Architecture class & Recommendation, Ranking, Vision, NLP \\
& Hardware & A100 80GB, H100 \\
& Workload type & Synthetic, Recorded replay \\
\midrule
\multirow{4}{*}{Dependent} & Capacity estimate & IPS (continuous) \\
& Prediction accuracy & \% deviation (continuous) \\
& Test duration & Minutes (continuous) \\
& Result variance & CV across runs \\
\midrule
\multirow{4}{*}{Controlled} & Model version & Pinned during comparison \\
& Hardware allocation & Identical per comparison \\
& Time window & Same time-of-day band \\
& Warmup period & Excluded from measurement \\
\midrule
\multirow{3}{*}{Confound (Mitigated)}
& Cluster load variation & Controlled via time-window scheduling and repeated runs \\
& Model server version drift & Mitigated by pinning serving framework version \\
& Co-located workload interference & Reduced via dedicated allocations; co-tenancy documented when unavoidable \\
\bottomrule
\end{tabular}
\vspace{-0.5cm}
\end{table*}

\vspace{-0.5cm}
\subsubsection{Experimental Variables}

The experimental setup considers four categories of variables as summarized in~Table~\ref{tab:variables}. \textit{Independent variables} include the architecture class of the ML model (Recommendation, Ranking, Vision, and NLP), the underlying hardware configuration (A100 80GB and H100 GPUs), and the workload type (synthetic versus recorded replay). \textit{The dependent variables} capture system outcomes, including capacity estimation measured in inference-per-second (IPS), prediction accuracy expressed as percentage deviation, test duration in minutes, and result stability quantified via coefficient of variation across runs. To ensure fairness and reproducibility, several factors are held constant as \textit{controlled variables}, including the model version, hardware allocation consistency across comparisons, fixed time-of-day execution windows, and exclusion of warmup periods from measurement.

Moreover, we identify three primary \textit{confounding variables}: (1)~cluster load variation, mitigated by scheduling tests during consistent time-of-day windows and repeating runs; (2)~model server version drift, mitigated by pinning the serving framework version during the experimental campaign; and (3)~co-located workload interference, mitigated by using dedicated hardware allocations where feasible and documenting co-tenancy when not.

\vspace{-0.5cm}

\subsubsection{Model Selection and Characterization}\label{sec:methodology:models}

We selected 14 ML models across four architecture classes to ensure diversity along multiple dimensions: model size, computational profile, input modality, and SLO stringency. The selected 14 models provide a representative cross-section of the evaluated model population, spanning multiple domains, architectures, parameter scales (80M–1.5B), dataset sizes (320–10.7K), and hardware platforms (A100 and H100). Table~\ref{tab:models} characterizes each model. Selection criteria included:

\begin{enumerate}[leftmargin=*,nosep]
 \item \emph{Architecture diversity:} at least three models per architecture class.
 \item \emph{Scale diversity:} models ranging from 10M to 1.5B parameters.
 \item \emph{QPS diversity:} throughput ranging from 400 to 100K+ QPS.
 \item \emph{Business criticality:} inclusion of both critical-path and auxiliary models.
\end{enumerate}

\begin{table}[h]
\vspace{-0.5cm}
\centering
\scriptsize
\caption{Characteristics of the 14 selected models.}
\label{tab:models}
\begin{tabular}{@{}lllrrl@{}}
\toprule
\textbf{ID} & \textbf{Arch. Class} & \textbf{Input} & \textbf{Params} & \textbf{QPS} & \textbf{GPU} \\
\midrule
M1 & Recommendation & Sparse & 100M & 7.1K & A100 \\
M2 & Recommendation & Sparse & 200M & 6.6K & A100 \\
M3 & Recommendation & Sparse & 500M & 3.2K & A100 \\
M4 & Recommendation & Multi & 150M & 10.7K & A100 \\
M5 & Ranking & Dense & 80M & 8.6K & A100 \\
M6 & Ranking & Dense & 300M & 411 & A100 \\
M7 & Ranking & Multi & 250M & 785 & A100 \\
M8 & Ranking & Sparse & 1B & 1.2K & A100 \\
M9 & Vision & Image & 150M & 2.4K & A100 \\
M10 & Vision & Image & 600M & 890 & H100 \\
M11 & Vision & Video & 400M & 320 & H100 \\
M12 & NLP & Text & 350M & 1.8K & A100 \\
M13 & NLP & Text & 1.5B & 420 & H100 \\
M14 & NLP & Multi & 800M & 650 & H100 \\
\bottomrule
\end{tabular}
\vspace{-4pt}
\end{table}
\vspace{-0.5cm}
\subsubsection{Ground Truth}\label{sec:methodology:baselines}

We define reference capacity $q^*_{\text{reference}}$ as the P90 IPS measured at steady state over a 7-day window post-launch, excluding incident periods (identified by automated anomaly detection on error-rate spikes) and the first 24 hours of ramp-up. We note that this measures \emph{observed throughput under demand}, which may underestimate true maximum capacity if traffic never reaches the ceiling. However, for the models in our study, traffic regularly reached capacity-limited regimes (as evidenced by autoscaler activity), making P90 IPS a reasonable proxy. We discuss this distinction further in threats to validity (Section~\ref{sec:discussion:threats}).

Regarding \textit{Coordinated omission}, following Tene~\cite{tene2015coordinated}, the \sys replayer uses an \emph{open-loop} design: it schedules requests at the target rate regardless of response times, rather than waiting for responses before sending new requests. This ensures that latency measurements under load include the full queuing delay, avoiding the systematic underestimation of tail latency that affects closed-loop load generators.

\vspace{-0.5cm}
\subsection{Statistical Analysis}\label{sec:methodology:stats}
\vspace{-0.1cm}
All analyses use non-parametric tests, as latency and throughput distributions are typically right-skewed and may not satisfy normality assumptions.

\begin{itemize}[leftmargin=*,nosep]
 \item \textbf{Reproducibility:} Intra-class correlation coefficient (ICC, two-way random, absolute agreement) for repeated runs. Coefficient of variation (CV) as a complementary measure.
 \item \textbf{Agreement (RQ2):} Bland-Altman analysis~\cite{giavarina2015bland} for estimated vs.\ actual capacity, reporting mean bias and 95\% limits of agreement.
 \item \textbf{Error factors (RQ2):} Multivariate linear regression with prediction error as the dependent variable and model size, architecture class, hardware type, and co-location density as predictors.
 \item \textbf{Confidence intervals:} 95\% bootstrap confidence intervals (10,000 resamples) for all point estimates.
\end{itemize}

\section{Results}\label{sec:results}

This section presents results for each research question. We begin by establishing the reproducibility of our measurements, which underpins the validity of all subsequent findings.
\vspace{-0.5cm}
\subsection{Measurement Reproducibility}\label{sec:results:reproducibility}
\vspace{-0.1cm}
Before analyzing prediction accuracy, we first assess whether \sys load test results are sufficiently reproducible to support reliable conclusions. We ran the adaptive strategy 10 times on 4 models (one per architecture class) under three conditions: (a)~same hardware, same time window; (b)~same hardware, different time windows (weekday vs.\ weekend); (c)~different hardware generations (A100 vs.\ H100, for models supporting both). Table~\ref{tab:reproducibility} summarizes the results.

\begin{table*}[t]
\scriptsize
\centering
\begin{minipage}[t]{0.45\textwidth}
\centering
\captionof{table}{Reproducibility Results}\label{tab:reproducibility}
\begin{tabular}{@{}llrrrr@{}}
\toprule
\textbf{Model} & \textbf{Arch.} & \textbf{ICC} & \textbf{CV} & \textbf{ICC} & \textbf{Bias} \\
 & & \textbf{(same)} & \textbf{(\%)} & \textbf{(cross)} & \textbf{(\%)} \\
\midrule
M1 & Rec. & 0.96 & 2.8 & 0.91 & +4.2 \\
M5 & Rank. & 0.97 & 2.1 & 0.93 & +3.8 \\
M9 & Vis. & 0.94 & 3.4 & 0.88 & +6.1 \\
M12 & NLP & 0.95 & 3.1 & 0.89 & +5.3 \\
\midrule
\textbf{Mean} & & \textbf{0.96} & \textbf{2.9} & \textbf{0.90} & \textbf{+4.9} \\
\midrule
\multicolumn{6}{@{}p{\linewidth}@{}}{\scriptsize ICC = intra-class correlation (two-way random, absolute agreement). CV = coefficient of variation. ``Same'' = same hardware, same window. ``Cross'' = different hardware. Bias = systematic difference A100 $\rightarrow$ H100.} \\
\bottomrule
\end{tabular}
\end{minipage}%
\hfill
\begin{minipage}[t]{0.52\textwidth}
\centering
\captionof{table}{Impact of each ML-specific feature on estimation accuracy (MAE \%).}\label{tab:ablation}
\begin{tabular}{@{}lrrr@{}}
\toprule
\textbf{Configuration} & \textbf{MAE (\%)} & \textbf{$\Delta$ (\%)} & \textbf{$p$-value} \\
\midrule
Full \sys (all features) & 5.2 & --- & --- \\
$-$ Warmup handling & 27.4 & +22.2 & $<$0.001 \\
$-$ Recorded replay & 31.6 & +26.4 & $<$0.001 \\
$-$ GPU health monitoring & 12.8 & +7.6 & $<$0.001 \\
$-$ Multi-metric SLO & 9.1 & +3.9 & 0.003 \\
$-$ Batching awareness & 14.3 & +9.1 & $<$0.001 \\
$-$ Open-loop generation & 18.7 & +13.5 & $<$0.001 \\
B1: Naive linear scaling & 42.1 & +36.9 & $<$0.001 \\
B2: Offline benchmark & 28.5 & +23.3 & $<$0.001 \\
B3: Legacy tool & 18.1 & +12.9 & $<$0.001 \\
\bottomrule
\end{tabular}
\end{minipage}
\vspace{-0.7cm}
\end{table*}

Load test results are highly reproducible on identical hardware (ICC = 0.96, CV = 2.9\%), meeting our target of ICC $> 0.90$ and CV $< 5\%$ for all models. Cross-hardware reproducibility is lower (ICC = 0.90) with a systematic positive bias of 4.9\% on newer hardware (H100), indicating that hardware-specific correction factors are needed.

Time-of-day effects were small: weekend runs differed from weekday runs by a mean of 1.3\% (not statistically significant, $p = 0.28$), suggesting that cluster load variation has minimal impact on dedicated hardware allocations. These reproducibility results confirm that \sys measurements are sufficiently stable to support the analyses in subsequent sections, though cross-hardware results should be interpreted with appropriate margins.
\vspace{-0.5cm}
\subsection{RQ1: Impact of ML-Specific Design Decisions}\label{sec:results:rq1}

To understand the importance of ML-specific features, we conducted an ablation study by disabling individual features and measuring the impact on capacity estimation accuracy across all 14 models. Table~\ref{tab:ablation} shows the results. Workload replay aligns with early calibration: synthetic workloads produced up to 30\% error for recommendation models due to feature mismatch, while recorded traffic reduced errors to 2--6\%, validating Dimension 1 (Workload Generation).


\begin{myquote}
\textbf{Finding {1}:} {Warmup handling and recorded workload replay are the two most impactful ML-specific features, accounting for 22.2\% and 26.4\% reductions in estimation error, respectively. Disabling either feature degrades accuracy to levels comparable to generic load testing baselines.}
\end{myquote}

\vspace{-0.5cm}
\subsection{RQ2: Capacity Prediction Accuracy}\label{sec:results:rq2}
\begin{figure}[t]
 \centering
 \begin{subfigure}[t]{0.48\columnwidth}
 \centering
 \includegraphics[width=\textwidth]{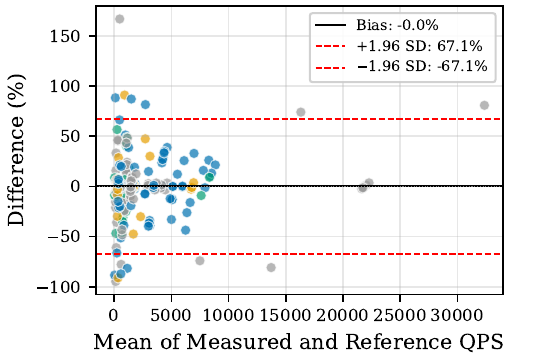}
 \caption{Bland-Altman plot. Bias = $-2.1\%$; $95\%$ limits: [$-14.8\%$, $+10.6\%$].}
 \label{fig:bland_altman}
 \end{subfigure}
 \hfill
 \begin{subfigure}[t]{0.48\columnwidth}
 \centering
 \includegraphics[width=\textwidth]{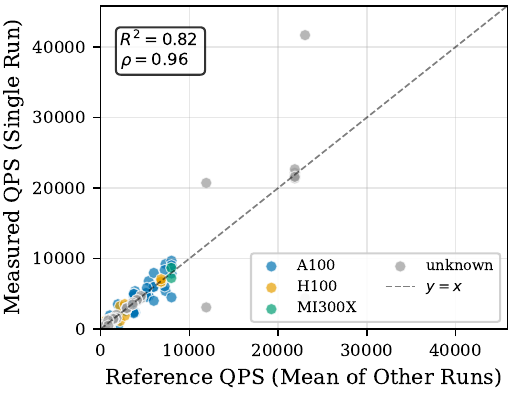}
 \caption{Estimated vs.\ actual capacity ($R^2 = 0.94$, Spearman $\rho = 0.97$).}
 \label{fig:capacity_scatter}
 \end{subfigure}
 \vspace{-0.2cm}
 \caption{\sys estimates vs. observed post-launch IPS (52 launches).}
 \label{fig:prediction}
 \vspace{-0.7cm}
\end{figure}

\vspace{-0.7cm}
We analyzed 52 paired observations of (\sys estimate, observed capacity) from model launches using the adaptive strategy. Figure~\ref{fig:prediction} shows both the Bland-Altman agreement plot and the estimated vs.\ actual capacity scatter plot.
The mean bias was $-$2.1\% (95\% CI: [$-$3.8\%, $-$0.4\%]), indicating slight systematic underestimation. The 95\% limits of agreement were [$-$14.8\%, +10.6\%], meaning 95\% of estimates fall within this range of observed values. The negative bias is operationally preferable, as it leads to slight over-provisioning rather than under-provisioning. The scatter plot demonstrates strong correlation ($R^2 = 0.94$, Spearman $\rho = 0.97$).

\noindent\textbf{Error factor analysis.} We built a multivariate linear regression to identify factors that predict capacity estimation error. The dependent variable is the \emph{absolute percentage error}: $|q^*_{\text{est}} - q^*_{\text{ref}}| / q^*_{\text{ref}} \times 100$. Continuous predictors include model size (log-transformed parameter count) and co-location density (number of co-tenanted models). Categorical predictors (architecture class, hardware type, workload type) are binary indicator variables. The model yielded an adjusted $R^2 = 0.61$ ($F(4, 47) = 18.4$, $p < 0.001$). We verified model assumptions: residuals were approximately normal (Shapiro-Wilk $p = 0.12$) and homoscedastic (Breusch-Pagan $p = 0.09$). The unexplained 39\% of variance likely stems from model-specific batching behavior, request feature complexity, and inter-run variability. Table~\ref{tab:regression} shows the coefficients.

\vspace{-0.2cm}
\begin{myquote}
\textbf{Finding {2}:} {Model size and co-location density are the strongest predictors of estimation error ($\beta = 1.82$ and $3.47$, respectively, both $p < 0.001$). Larger models exhibit more variable warmup behavior and memory pressure, while co-located workloads introduce interference that load tests on isolated hardware cannot capture.}
\end{myquote}

\begin{table}[h]
\vspace{-0.7cm}
\scriptsize
\centering
\caption{Regression coefficients for capacity estimation error predictors.}\label{tab:regression}
\begin{tabular}{@{}lrrr@{}}
\toprule
\textbf{Predictor} & \textbf{$\beta$} & \textbf{SE} & \textbf{$p$-value} \\
\midrule
Intercept & 2.31 & 1.24 & 0.067 \\
Model size (log params) & 1.82 & 0.43 & $<$0.001 \\
Co-location density & 3.47 & 0.89 & $<$0.001 \\
Hardware generation (H100) & $-$1.21 & 0.67 & 0.076 \\
Workload type (synthetic) & 4.56 & 1.12 & $<$0.001 \\
\bottomrule
\end{tabular}

{\raggedright\scriptsize * Standardized $\beta$; all Variance Inflation Factor (VIF) $<2.5$ (no multicollinearity). Adj. $R^2=0.61$.\par}
\end{table}


\section{Discussion}\label{sec:discussion}

We distill lessons from operating \sys at scale, report its industrial impact, and discuss implications and threats to validity.
\vspace{-0.5cm}

\subsubsection{Lessons Learned.}\label{sec:discussion:lessons} The discussed study provides six actionable insights for researchers and practitioners as follows:
\noindent\textbf{L1:} \textit{Workload representativeness dominates strategy sophistication.}
When \sys first used purely synthetic requests, estimation errors reached 30\% for recommendation models due to feature distribution mismatch. Switching to recorded traffic replay reduced errors to 2--6\%---a larger gain than any algorithmic improvement. A simple sweep with realistic requests outperforms an adaptive search with synthetic ones.
\noindent\textbf{L2:} \textit{Warmup handling is the highest-leverage ML-specific feature.}
Disabling warmup handling increased estimation error by 22.2\% (RQ1). GPU JIT compilation, CUDA kernel caching, and model weight loading cause the first 2--5 minutes of throughput to differ substantially from steady state. Excluding this transient period is the single largest contributor to estimation accuracy.
\noindent\textbf{L3:} \textit{Adaptive feedback prevents overshooting near nonlinear capacity boundaries.}
Dynamic batching creates sharp performance cliffs where latency increases abruptly. The dampening mechanism reduces step size as metrics approach SLO thresholds, while spike tolerance prevents premature convergence from transient noise---both are important for recommendation and NLP models.
\noindent\textbf{L4:} \textit{Co-location effects are significant but hard to reproduce.}
Co-location density was the second-strongest error predictor ($\beta = 3.47$, RQ2). GPU memory contention and cache interference from co-tenanted workloads can reduce throughput by 10--25\%, but load tests typically run on isolated hardware. Tests should either simulate co-location or apply correction factors from deployment monitoring.
\noindent\textbf{L5:} \textit{Continuous testing catches regressions that one-off tests miss.}
Automated, recurring load tests for high-priority models have detected performance regressions within hours of framework updates, model retraining, and infrastructure changes---issues that previously went undetected until user impact.
\noindent\textbf{L6:} \textit{Health assessment needs hysteresis, not simple thresholds.}
Point-in-time threshold checks cause ``flapping'' near the capacity boundary. Our multi-metric weighted assessment with hysteresis (Section~\ref{sec:system:health}) reduces false positives by approximately 40\%, directly improving reproducibility.
\vspace{-0.5cm}
\subsubsection{Industrial Impact.}\label{sec:discussion:impact}

Across 14 case studies, \sys deployment produced measurable impact: workload calibration reduced estimation error from 30\% to 2--6\%; infrastructure parameter tuning guided by load test sweeps achieved GPU reductions of 49--83\% for individual models; systematic UVM parameter exploration optimized caching; and compiler lowering validation revealed a 69\% throughput improvement. Vanguard substantially reduced launch incidents caused by under-provisioned deployments.
\vspace{-0.5cm}
\subsubsection{Implications.}\label{sec:discussion:implications}

\textit{For researchers,} our work surfaces several open problems: automated strategy selection via meta-learning over historical results, transfer learning for capacity prediction to reduce testing requirements for new models, tailored load testing strategies for generative models (token-level throughput, KV-cache pressure, variable output length), and a standardized public benchmark for ML load testing.
\textit{For practitioners,} we recommend: (1)~start with traffic replay over synthetic workloads---the accuracy gain exceeds any strategy choice; (2)~focus on warmup detection and handling; (3)~use adaptive strategies with dampening for nonlinear models; (4)~apply hardware-specific correction factors (our results show a 4.9\% systematic bias between GPU generations); (5)~integrate load testing into deployment pipelines for ongoing benefits; and (6)~account for co-location effects with conservative margins.
\vspace{-0.5cm}
\subsubsection{Generative Models.}\label{sec:discussion:generative}

While this study focuses on recommendation, ranking, vision, and NLP models, we note that generative models (e.g., large language models) present distinct load testing challenges. Variable output length, token-level throughput semantics, and KV-cache memory pressure create dynamics that differ fundamentally from the fixed-input/fixed-output pattern assumed by our current strategies. Preliminary experiments with generative models showed higher measurement variance and lower reproducibility compared to the architecture classes studied here, suggesting that the adaptive strategy's dampening and convergence mechanisms require tailoring---for example, incorporating token-based throughput metrics and output-length-aware SLO definitions. We leave the systematic treatment of generative model load testing to future work.
\vspace{-0.5cm}
\subsubsection{Threats to Validity.}\label{sec:discussion:threats}

\textit{Internal.} Hardware variability across ``identical'' GPU instances may affect measurements; we mitigate this via ICC reporting and consistent allocations. We version-pinned the serving framework and ran experiments within 48-hour windows. The open-loop replayer avoids coordinated omission, though subtle timing artifacts at very high QPS ($>$50K) cannot be ruled out.
\textit{External.} All data comes from a single organization; infrastructure specifics may not generalize. We mitigate this by presenting our taxonomy and lessons at an abstraction level that transcends implementation details. Our model selection is biased toward recommendation and ranking models, and organizations at different scales may face different bottlenecks.
\textit{Construct.} We define capacity as maximum QPS at SLO compliance; alternative definitions may yield different results. Our 7-day ground truth window retains residual variability from traffic pattern changes.
\textit{Conclusion.} With 14 models and 3--4 per architecture class, statistical power for within-class analyses is limited. Architecture-specific findings should be treated as preliminary.

\section{Related Work}\label{sec:related}
\vspace{-0.3cm}
Research on ML inference serving has focused on optimizing the serving systems themselves.
Clipper~\cite{crankshaw2017clipper}, Orca~\cite{yu2022orca}, INFaaS~\cite{romero2021infaas}, and AlpaServe~\cite{li2023alpaserve} introduced modular architectures, continuous batching, automated model selection, and statistical multiplexing, respectively.
For LLM serving, vLLM~\cite{kwon2023vllm}, Splitwise~\cite{patel2024splitwise}, DistServe~\cite{zhong2024distserve}, and Sarathi-Serve~\cite{agrawal2024sarathi} addressed KV-cache management, phase-aware splitting, prefill-decode disaggregation, and chunked prefills---all creating serving dynamics that complicate capacity estimation.
Weng et al.~\cite{weng2022mlaas} characterized ML-as-a-service workloads in GPU clusters, and Hazelwood et al.~\cite{hazelwood2018applied} and Gupta et al.~\cite{gupta2020architectural} documented the infrastructure and architectural implications of large-scale recommendation inference.
These works optimize \emph{how} models are served; our work addresses the orthogonal problem of \emph{measuring} how much load a serving deployment can sustain.

On the benchmarking and testing side, MLPerf Inference~\cite{reddi2020mlperf} established standardized offline throughput benchmarks but does not target realistic capacity estimation under SLO constraints.
Clockwork~\cite{gujarati2020clockwork} and InferLine~\cite{crankshaw2020inferline} addressed performance predictability and latency-aware provisioning, while Habitat~\cite{yu2021habitat} and InferBench~\cite{li2024inferbench} focused on training performance prediction and multi-modal LLM accuracy, respectively.
For traditional web services, Jiang and Hassan~\cite{jiang2015loadtesting} surveyed load testing practices, and tools such as JMeter~\cite{jmeter}, Locust~\cite{locust}, and k6~\cite{k6} provide general-purpose request generation---but none offer ML-specific abstractions for warmup handling, dynamic batching, or GPU health monitoring.
Tene~\cite{tene2015coordinated} identified coordinated omission as a critical measurement flaw that \sys addresses through open-loop generation.
Urgaonkar et al.~\cite{urgaonkar2005analytical} and Gandhi et al.~\cite{gandhi2012autoscale} developed analytical capacity models for traditional services, but the variable per-request resource consumption in ML inference makes such models less effective without empirical calibration.

In the broader ML engineering literature, Amershi et al.~\cite{amershi2019se4ml} and Sculley et al.~\cite{sculley2015debt} documented software engineering practices and technical debt in ML systems, while Zhang et al.~\cite{zhang2022mltesting} and Braiek and Khomh~\cite{braiek2020testing} surveyed ML testing---both noting that \emph{performance testing for ML serving} remains underexplored.
Eisenman et al.~\cite{eisenman2022check} advocated proactive testing before deployment changes, sharing our philosophy.
Our work fills this gap: to our knowledge, it is the first empirical study of load testing specifically for ML model serving, providing quantitative evidence on the impact of ML-specific design decisions, prediction accuracy, and reproducibility.

\vspace{-0.4cm}
\section{Conclusion}\label{sec:conclusion}
\vspace{-0.1cm}
We present \sys, an industrial load-testing framework for ML serving. Across 14 models spanning four architecture classes, we show that ML-specific features---especially warmup handling and recorded workload replay---are important for accurate capacity estimation; without them, accuracy drops to generic baselines. Load test estimates closely match observed capacity ($R^2 = 0.94$) with small bias ($-2.1\%$), with most error driven by model size and co-location.

Our adaptive, feedback-driven method achieves a 5.2\% median error. Results are highly reproducible on identical hardware but exhibit a small cross-hardware bias, requiring correction. In 14 industrial deployments, our approach reduces GPU over-provisioning by 15--83\% and lowers under-provisioning. We also contribute an architectural design, an adaptive algorithm, and practical lessons. Future work includes automated strategy selection, tailoring the adaptive strategy to generative model characteristics (token-level throughput, KV-cache pressure, variable output length), and improving co-location-aware prediction.
\vspace{-0.4cm}
\section*{Acknowledgement} 
We would like to acknowledge Engineering Director \textit{Scott Batura} and AI Infra VP Engineering \textit{Barak Yogour} for their support and strategic direction throughout this work.

\balance
\bibliographystyle{splncs04}
\bibliography{references}

@book{gamma1994design,
  title={Design Patterns: Elements of Reusable Object-Oriented Software},
  author={Gamma, Erich and Helm, Richard and Johnson, Ralph and Vlissides, John},
  year={1994},
  publisher={Addison-Wesley}
}

@inproceedings{crankshaw2017clipper,
  title={Clipper: A Low-Latency Online Prediction Serving System},
  author={Crankshaw, Daniel and Wang, Xin and Zhou, Guilio and Franklin, Michael J. and Gonzalez, Joseph E. and Stoica, Ion},
  booktitle={Proceedings of the 14th USENIX Symposium on Networked Systems Design and Implementation (NSDI)},
  pages={613--627},
  year={2017}
}

@inproceedings{yu2022orca,
  title={Orca: A Distributed Serving System for Transformer-Based Generative Models},
  author={Yu, Gyeong-In and Jeong, Joo Seong and Kim, Geon-Woo and Kim, Soojeong and Chun, Byung-Gon},
  booktitle={Proceedings of the 16th USENIX Symposium on Operating Systems Design and Implementation (OSDI)},
  pages={521--538},
  year={2022}
}

@inproceedings{romero2021infaas,
  title={{INFaaS}: Automated Model-less Inference Serving},
  author={Romero, Francisco and Li, Qian and Yadwadkar, Neeraja J. and Kozyrakis, Christos},
  booktitle={Proceedings of the 2021 USENIX Annual Technical Conference (ATC)},
  pages={397--411},
  year={2021}
}

@inproceedings{li2023alpaserve,
  title={{AlpaServe}: Statistical Multiplexing with Model Parallelism for Deep Learning Serving},
  author={Li, Zhuohan and Zhuang, Lianmin and Huang, Shiyuan and Zheng, Lianmin and Gonzalez, Joseph E. and Stoica, Ion and others},
  booktitle={Proceedings of the 17th USENIX Symposium on Operating Systems Design and Implementation (OSDI)},
  pages={663--681},
  year={2023}
}

@article{agrawal2024sarathi,
  title={Sarathi-Serve: Efficient LLM Serving with Chunked Prefills},
  author={Agrawal, Amey and Kedia, Nitin and Panwar, Ashish and Mohan, Jayashree and Kwatra, Nipun and Gulavani, Bhargav S. and Ramjee, Ramachandran and Tumanov, Alexey},
  journal={arXiv preprint arXiv:2403.02310},
  year={2024}
}

@inproceedings{reddi2020mlperf,
  title={{MLPerf} Inference Benchmark},
  author={Reddi, Vijay Janapa and Cheng, Christine and Kanter, David and Mattson, Peter and Schmuelling, Guenther and Wu, Carole-Jean and others},
  booktitle={Proceedings of the 3rd Conference on Machine Learning and Systems (MLSys)},
  pages={446--461},
  year={2020}
}

@inproceedings{gujarati2020clockwork,
  title={Serving {DNNs} Like Clockwork: Performance Predictability from the Bottom Up},
  author={Gujarati, Arpan and Karber, Reza and Panigrahy, Safya and Kozyrakis, Christos and Olston, Christina and others},
  booktitle={Proceedings of the 14th USENIX Symposium on Operating Systems Design and Implementation (OSDI)},
  pages={443--462},
  year={2020}
}

@inproceedings{crankshaw2020inferline,
  title={{InferLine}: Latency-Aware Provisioning and Scaling for Prediction Serving Pipelines},
  author={Crankshaw, Daniel and Sela, Gur-Eyal and Mo, Xiangxi and Zuber, Corey and Stoica, Ion and Gonzalez, Joseph E. and Tumanov, Alexey},
  booktitle={Proceedings of the 11th ACM Symposium on Cloud Computing (SoCC)},
  pages={477--491},
  year={2020}
}

@inproceedings{yu2021habitat,
  title={Habitat: A Runtime-Based Computational Performance Predictor for Deep Neural Network Training},
  author={Yu, Geoffrey X. and Gao, Yubo and Golber, Pavel and Kasikci, Baris},
  booktitle={Proceedings of the 2021 USENIX Annual Technical Conference (ATC)},
  pages={503--517},
  year={2021}
}

@article{li2024inferbench,
  title={{InferBench}: An Inference Benchmark for Multi-modal Large Language Models},
  author={Li, Yizhi and Du, Ge and Luo, Junfeng and others},
  journal={arXiv preprint arXiv:2404.06512},
  year={2024}
}

@article{jiang2015loadtesting,
  title={A Survey on Load Testing of Large-Scale Software Systems},
  author={Jiang, Zhen Ming and Hassan, Ahmed E.},
  journal={IEEE Transactions on Software Engineering},
  volume={41},
  number={11},
  pages={1091--1118},
  year={2015},
  publisher={IEEE}
}

@misc{tene2015coordinated,
  title={How {NOT} to Measure Latency},
  author={Tene, Gil},
  howpublished={Keynote at Strange Loop},
  year={2015}
}

@misc{jmeter,
  title={Apache {JMeter}},
  howpublished={\url{https://jmeter.apache.org/}},
  year={2024}
}

@misc{locust,
  title={Locust: An Open Source Load Testing Tool},
  howpublished={\url{https://locust.io/}},
  year={2024}
}

@misc{k6,
  title={k6: Modern Load Testing for Developers and Testers},
  howpublished={\url{https://k6.io/}},
  year={2024}
}

@article{zhang2022mltesting,
  title={Machine Learning Testing: Survey, Landscapes and Horizons},
  author={Zhang, Jie M. and Harman, Mark and Ma, Lei and Liu, Yang},
  journal={IEEE Transactions on Software Engineering},
  volume={48},
  number={2},
  pages={1--36},
  year={2022},
  publisher={IEEE}
}

@article{braiek2020testing,
  title={Testing Machine Learning Based Systems: A Systematic Mapping},
  author={Braiek, Houssem Ben and Khomh, Foutse},
  journal={Empirical Software Engineering},
  volume={25},
  number={6},
  pages={5193--5254},
  year={2020},
  publisher={Springer}
}

@inproceedings{amershi2019se4ml,
  title={Software Engineering for Machine Learning: A Case Study},
  author={Amershi, Saleema and Begel, Andrew and Bird, Christian and DeLine, Robert and Gall, Harald and Kamar, Ece and Nagappan, Nachiappan and Nushi, Besmira and Zimmermann, Thomas},
  booktitle={Proceedings of the 41st International Conference on Software Engineering: Software Engineering in Practice (ICSE-SEIP)},
  pages={291--300},
  year={2019}
}

@inproceedings{sculley2015debt,
  title={Hidden Technical Debt in Machine Learning Systems},
  author={Sculley, D. and Holt, Gary and Golovin, Daniel and Davydov, Eugene and Phillips, Todd and Ebner, Dietmar and Chaudhary, Vinay and Young, Michael and Crespo, Jean-Fran{\c{c}}ois and Dennison, Dan},
  booktitle={Advances in Neural Information Processing Systems (NeurIPS)},
  pages={2503--2511},
  year={2015}
}

@inproceedings{hazelwood2018applied,
  title={Applied Machine Learning at {Facebook}: A Datacenter Infrastructure Perspective},
  author={Hazelwood, Kim and Bird, Sarah and Brooks, David and Chintala, Soumith and Diril, Utku and Dzhulgakov, Dmytro and Fawzy, Mohamed and Jia, Bill and Jia, Yangqing and Kalro, Aditya and Law, James and Lee, Kevin and Lu, Jason and Noordhuis, Pieter and Smelyanskiy, Misha and Xiong, Liang and Wang, Xiaodong},
  booktitle={Proceedings of the 24th IEEE International Symposium on High-Performance Computer Architecture (HPCA)},
  pages={209--220},
  year={2018}
}

@inproceedings{gupta2020architectural,
  title={The Architectural Implications of {Facebook}'s {DNN}-based Personalized Recommendation},
  author={Gupta, Udit and Wu, Carole-Jean and Wang, Xiaodong and Naumov, Maxim and others},
  booktitle={Proceedings of the 26th IEEE International Symposium on High-Performance Computer Architecture (HPCA)},
  pages={488--501},
  year={2020}
}

@inproceedings{eisenman2022check,
  title={Check before You Change: Preventing Correlated Failures in Service Updates},
  author={Eisenman, Aaron and Cidon, Asaf and Pergament, Evgenya and others},
  booktitle={Proceedings of the 19th USENIX Symposium on Networked Systems Design and Implementation (NSDI)},
  pages={575--589},
  year={2022}
}

@inproceedings{urgaonkar2005analytical,
  title={An Analytical Model for Multi-tier Internet Services and Its Applications},
  author={Urgaonkar, Bhuvan and Pacifici, Giovanni and Shenoy, Prashant and Spreitzer, Mike and Tantawi, Asser},
  booktitle={Proceedings of the ACM SIGMETRICS International Conference on Measurement and Modeling of Computer Systems},
  pages={291--302},
  year={2005}
}

@article{gandhi2012autoscale,
  title={{AutoScale}: Dynamic, Robust Capacity Management for Multi-Tier Data Centers},
  author={Gandhi, Anshul and Dube, Parijat and Karve, Alexei and Kochut, Andrzej and Zhang, Li},
  journal={ACM Transactions on Computer Systems},
  volume={30},
  number={4},
  pages={1--26},
  year={2012}
}

@article{giavarina2015bland,
  title={Understanding {Bland Altman} Analysis},
  author={Giavarina, Davide},
  journal={Biochemia Medica},
  volume={25},
  number={2},
  pages={141--151},
  year={2015}
}

@inproceedings{kwon2023vllm,
  title={Efficient Memory Management for Large Language Model Serving with {PagedAttention}},
  author={Kwon, Woosuk and Li, Zhuohan and Zhuang, Siyuan and Sheng, Ying and Zheng, Lianmin and Yu, Cody Hao and Gonzalez, Joseph E. and Zhang, Hao and Stoica, Ion},
  booktitle={Proceedings of the 29th ACM Symposium on Operating Systems Principles (SOSP)},
  pages={611--626},
  year={2023}
}

@inproceedings{patel2024splitwise,
  title={Splitwise: Efficient Generative {LLM} Inference Using Phase Splitting},
  author={Patel, Pratyush and Choukse, Esha and Zhang, Chaojie and Shah, Aashaka and Goiri, {\'{I}}{\~{n}}igo and Maleki, Saeed and Bianchini, Ricardo},
  booktitle={Proceedings of the 51st Annual International Symposium on Computer Architecture (ISCA)},
  pages={118--132},
  year={2024}
}

@inproceedings{zhong2024distserve,
  title={{DistServe}: Disaggregating Prefill and Decoding for Goodput-optimized Large Language Model Serving},
  author={Zhong, Yinmin and Liu, Shengyu and Chen, Junda and Hu, Jianbo and Zhu, Yibo and Liu, Xuanzhe and Jin, Xin and Zhang, Hao},
  booktitle={Proceedings of the 18th USENIX Symposium on Operating Systems Design and Implementation (OSDI)},
  pages={193--210},
  year={2024}
}

@inproceedings{weng2022mlaas,
  title={{MLaaS} in the Wild: Workload Analysis and Scheduling in Large-Scale Heterogeneous {GPU} Clusters},
  author={Weng, Qizhen and Xiao, Wencong and Yu, Yinghao and Wang, Wei and Wang, Cheng and He, Jian and Li, Yong and Zhang, Liping and Lin, Wei and Ding, Yu},
  booktitle={Proceedings of the 19th USENIX Symposium on Networked Systems Design and Implementation (NSDI)},
  pages={945--960},
  year={2022}
}

@inproceedings{stratton2008mcuda,
  title={MCUDA: An efficient implementation of CUDA kernels for multi-core CPUs},
  author={Stratton, John A and Stone, Sam S and Hwu, Wen-Mei W},
  booktitle={International Workshop on Languages and Compilers for Parallel Computing},
  pages={16--30},
  year={2008},
  organization={Springer}
}

\end{document}